\documentclass{article}
\usepackage{iclr2020_conference,times}
\iclrfinalcopy

\usepackage{amsmath,amsfonts,bm}

\def\eqref#1{equation~\ref{#1}}

\def\1{\bm{1}}

\DeclareMathAlphabet{\mathsfit}{\encodingdefault}{\sfdefault}{m}{sl}
\SetMathAlphabet{\mathsfit}{bold}{\encodingdefault}{\sfdefault}{bx}{n}

\usepackage{url}
\usepackage{algorithm}
\usepackage{algpseudocode}
\usepackage{xcolor}
\usepackage{caption}
\usepackage{booktabs}

\usepackage{pgfplotstable}
\usetikzlibrary{pgfplots.groupplots}

\pgfplotsset{compat=1.14}
\pgfplotsset{
    tick label style = {font = \tiny},
    label style = {font = \small},
    legend style = {font = \footnotesize},
    every axis plot/.append style = {font = \scriptsize}
}

\title{Overlearning Reveals Sensitive Attributes}

\author{Congzheng Song \\
Cornell University\\
\texttt{cs2296@cornell.edu} \\
\And
Vitaly Shmatikov \\
Cornell Tech \\
\texttt{shmat@cs.cornell.edu} \\
}

\newcommand{\enc}{E}
\newcommand{\dec}{R}
\newcommand{\disc}{D}
\newcommand{\clf}{C}
\newcommand{\data}{\mathcal{D}}
\newcommand{\dataa}{\mathcal{D}_\text{aux}}
\newcommand{\datat}{\mathcal{D}_\text{transfer}}

\newcommand{\Maux}{M_\text{aux}}
\newcommand{\Mattack}{M_\text{attack}}
\newcommand{\Mtransfer}{M_\text{transfer}}

\newcommand{\expectation}{\mathbb{E}}

\newcommand{\paraname}[1]{\noindent{\bf \em #1}}
\newcommand{\paragraphbe}[1]{\noindent{\bf \em #1}\hspace*{.3em}}

\begin{document}

\maketitle
\begin{abstract}

``Overlearning'' means that a model trained for a seemingly simple
objective implicitly learns to recognize attributes and concepts that are
(1) not part of the learning objective, and (2) sensitive from a privacy
or bias perspective.  For example, a binary gender classifier of facial
images also learns to recognize races\textemdash even races that are
not represented in the training data\textemdash and identities.

We demonstrate overlearning in several vision and NLP models and analyze
its harmful consequences.  First, inference-time representations of an
overlearned model reveal sensitive attributes of the input, breaking
privacy protections such as model partitioning.  Second, an overlearned
model can be ``re-purposed'' for a different, privacy-violating task
even in the absence of the original training data.

We show that overlearning is intrinsic for some tasks and cannot be
prevented by censoring unwanted attributes.  Finally, we investigate
where, when, and why overlearning happens during model training.



\end{abstract}


\section{Introduction}

We demonstrate that representations learned by deep models when training
for seemingly simple objectives reveal privacy- and bias-sensitive
attributes that are not part of the specified objective.  These
unintentionally learned concepts are neither finer-, nor coarse-grained
versions of the model's labels, nor statistically correlated with them.
We call this phenomenon \textbf{overlearning}.  For example, a binary
classifier trained to determine the gender of a facial image also learns
to recognize races (including races not represented in the training data)
and even identities of individuals.




Overlearning has two distinct consequences.  First, the model's
inference-time representation of an input reveals the input's sensitive
attributes.  For example, a facial recognition model's representation
of an image reveals if two specific individuals appear together in it.
Overlearning thus breaks inference-time privacy protections based on
model partitioning~\citep{Osia2018Deep, Chi2018Privacy, Wang2018Not}.
Second, we develop a new, transfer learning-based technique to
``re-purpose'' a model trained for benign task into a model for a
different, privacy-violating task.  This shows the inadequacy of privacy
regulations that rely on explicit enumeration of learned attributes.


Overlearning is intrinsic for some tasks, i.e., it is not possible to
prevent a model from learning sensitive attributes.  We show that if these
attributes are censored~\citep{Xie2017ControllableIT, Moyer2018Invariant},
the censored models either fail to learn their specified tasks, or still
leak sensitive information.  We develop a new de-censoring technique to
extract information from censored representations.  We also show that
overlearned representations enable recognition of sensitive attributes
that are not present in the training data.  Such attributes cannot be
censored using any known technique.  This shows the the inadequacy of
censoring as a privacy protection technology.

To analyze where and why overlearning happens, we empirically show how
general features emerge in the lower layers of models trained for simple
objectives and conjecture an explanation based on the complexity of the
training data.
\section{Background}

We focus on supervised deep learning.  Given an input $x$, a model $M$
is trained to predict the target $y$ using a discriminative approach.
We represent the model $M=\clf \circ \enc$ as a feature extractor
(encoder) $\enc$ and classifier $\clf$.  The representation $z=\enc(x)$ is
passed to $\clf$ to produce the prediction by modeling $p(y|z)= \clf(z)$.
Since $\enc$ can have multiple layers of representation, we use $\enc_l(x)
= z_l$ to denote the model's internal representation at layer $l$; $z$
is the representation at the last layer.

\label{sec:partition}
\noindent
\textbf{\em Model partitioning} splits the model into a local,
on-device part and a remote, cloud-based part to improve scalability
of inference~\citep{Lane2015Can, Kang2017Neurosurgeon} and protect
privacy of inputs into the model~\citep{Li2017Privynet, Osia2018Deep,
Chi2018Privacy, Wang2018Not}.  For privacy, the local part of the model
computes a representation, censors it as described below, and sends it
to the cloud part, which computes the model's output.


\paragraphbe{Censoring representations.}
\label{sec:censor}
The goal is to encode input $x$ into a representation $z$ that does not
reveal unwanted properties of $x$, yet is expressive enough to predict the
task label $y$.  Censoring has been used to achieve transform-invariant
representations for computer vision, bias-free representations for
fair machine learning, and privacy-preserving representations that hide
sensitive attributes.

A straightforward censoring approach is based on \emph{adversarial
training}~\citep{Goodfellow2014Generative}.  It involves a
mini-max game between a discriminator $\disc$ trying to infer $s$
from $z$ during training and an encoder and classifier trying to
infer the task label $y$ while minimizing the discriminator's
success~\citep{Edwards2016CensoringRW, Iwasawa2017Privacy,
Hamm2017Minimax, Xie2017ControllableIT, Li2018Towards, Coavoux2018Privacy,
Elazar2018Adversarial}.  The game is formulated as:
\begin{align}
\min_{\enc, \clf} \max_{\disc} \expectation_{x, y, s}[\gamma \log p(s|z=\enc(x)) - \log p(y|z=\enc(x))]
\end{align}
\noindent
where $\gamma$ balances the two log likelihood terms.  The inner
optimization maximizes $\log p(s|z=\enc(x))$, i.e., the discriminator's
prediction of the sensitive attribute $s$ given a representation $z$.
The outer optimization, on the other hand, trains the encoder and
classifier to minimize the log likelihood of the discriminator predicting
$s$ and maximize that of predicting the task label $y$.


Another approach casts censoring as a single \emph{information-theoretical}
objective.  The requirement that $z$ not reveal $s$ can be formalized as
an independence constraint $z\perp s$, but independence is intractable
to measure in practice, thus the requirement is relaxed to a constraint
on the mutual information between $z$ and $s$~\citep{Osia2018Deep,
Moyer2018Invariant}.  The overall training objective of censoring $s$
and predicting $y$ from $z$ is formulated as:
\begin{align}
    \max I(z, y) - \beta I(z, x) - \lambda I(z, s)
\label{eq:itcensor}
\end{align}
\noindent
where $I$ is mutual information and $\beta, \lambda$ are the balancing
coefficients; $\beta=0$ in~\citep{Osia2018Deep}.  The first two terms
$I(z, y) - \beta I(z, x)$ is the objective of variational information
bottleneck~\citep{Alemi2017Deep}, the third term is the relaxed
independence constraint of $z$ and $s$.


Intuitively, this objective aims to maximize the information of $y$ in $z$
as per $I(z, y)$, forget the information of $x$ in $z$ as per $-\beta I(z,
x)$, and remove the information of $s$ in $z$ as per $-\lambda I(z, s)$.
This objective has an analytical lower bound~\citep{Moyer2018Invariant}:
\begin{align}
\expectation_{x, s}[\expectation_{z,y}[\log p(y|z)] - (\beta + \lambda)KL[q(z|x)||q(z)] - \lambda \expectation_z[\log p(x|z, s)]]
\end{align}
where $KL$ is Kullback-Leibler divergence and $\log p(x|z, s)$ is the
reconstruction likelihood of $x$ given $z$ and $s$.  The conditional
distributions $p(y|z)=\clf(z)$, $q(z|x)=\enc(x)$ are modeled as in
adversarial training and $p(x|z, s)$ is modeled with a decoder $\dec(z,
s)=p(x|z, s)$.


All known censoring techniques require a ``blacklist'' of attributes
to censor, and inputs with these attributes must be represented in
the training data.  Censoring for fairness is applied to the model's
final layer to make its output independent of the sensitive attributes
or satisfy a specific fairness constraint~\citep{Zemel2013Learning,
Louizos2015Variational, Madras2018Learning, Song2019Learning}.  In this
paper, we use censoring not for fairness but to demonstrate that models
cannot be prevented from learning to recognize sensitive attributes.
To show this, we apply censoring to different layers, not just the output.




\section{Exploiting Overlearning}

We demonstrate two different ways to exploit overlearning in a trained
model $M$.  The inference-time attack (Section~\ref{sec:infer-attack})
applies $M$ to an input and uses $M$'s representation of that input
to predict its sensitive attributes.  The model-repurposing attack
(Section~\ref{sec:transfer-attack}) uses $M$ to create another model that,
when applied to an input, directly predicts its sensitive attributes.

\begin{figure}[t]
\footnotesize
\noindent\fbox{
\begin{minipage}{0.46\linewidth}
\textbf{Inferring $s$ from representation:}
\begin{algorithmic}[1]
\State \textbf{Input:} Adversary's auxiliary dataset $\dataa$, black-box
oracle $\enc$, observed  $z^\star$
\State $\data_\text{attack} \gets \{(\enc(x), s) \, | \, (x, s)\in \dataa\}$
\State Train attack model $\Mattack$ on $\data_\text{attack}$ 
\State \Return prediction $\hat{s} = \Mattack(z^\star)$
\end{algorithmic}
\end{minipage}
}
\hfill
\noindent\fbox{
\begin{minipage}{0.46\linewidth}
\textbf{Adversarial re-purposing:}
\begin{algorithmic}[1]
\State \textbf{Input:} Model $M$ for the original task,
transfer dataset $\datat$ for the new task
\State 
Build $\Mtransfer = \clf_\text{transfer} \circ \enc_l$ on layer $l$
\State Fine-tune $\Mtransfer$ on $\datat$ 
\State \Return transfer model $\Mtransfer$
\end{algorithmic}
\end{minipage}
}
\label{fig:pseudocode}
\caption{Pseudo-code for inference from representation and adversarial re-purposing}
\end{figure}


\subsection{Inferring sensitive attributes from representation}
\label{sec:infer-attack}

We measure the leakage of sensitive properties from the representations
of overlearned models via the following attack.  Suppose an adversary
can observe the representation $z^\star$ of a trained model $M$ on input
$x^\star$ at inference time but cannot observe $x^\star$ directly.  This
scenario arises in practice when model evaluation is partitioned in order
to protect privacy of inputs\textemdash see Section~\ref{sec:partition}.
The adversary wants to infer some property $s$ of $x^\star$ that is not
part of the task label $y$.



We assume that the adversary has an auxiliary set $\dataa$ of labeled
$(x, s)$ pairs and black-box oracle $\enc$ to compute the corresponding
$\enc(x)$.  The purpose of $\dataa$ is to help the adversary recognize
the property of interest in the model's representations; it need not be
drawn from the same dataset as $x^{\star}$.  The adversary uses supervised
learning on the $(\enc(x), s)$ pairs to train an attack model $\Mattack$.
At inference time, the adversary predicts $\hat{s}$ from the observed
$z^\star$ as $\Mattack(z^\star)$.


\paragraphbe{De-censoring.}
\label{sec:decensor}
If the representation $z$ is ``censored'' (see Section~\ref{sec:censor})
to reduce the amount of information it reveals about $s$, the direct
inference attack may not succeed.  We develop a new, learning-based
\emph{de-censoring} approach (see Algorithm~\ref{algo:decensor}) to
convert censored representations into a different form that leaks more
information about the property of interest.  The adversary trains $\Maux$
on $\dataa$ to predict $s$ from $x$, then transforms $z$ into the input
features of $\Maux$.

We treat de-censoring as an optimization problem with a feature space
$L_2$ loss $||T(z) - z_\text{aux}||_2^2$, where $T$ is the transformer
that the adversary wants to learn and $z_\text{aux}$ is the uncensored
representation from $\Maux$.  Training with a feature-space loss has
been proposed for synthesizing more natural images by matching them with
real images~\citep{Dosovitskiy2016Generating, Nguyen2016Synthesizing}.
In our case, we match censored and uncensored representations.
The adversary can then use $T(z)$ as an uncensored approximation of
$z$ to train an inference model $\Mattack$ and infer property $s$
as $\Mattack(T(z^\star))$.

\begin{algorithm}[t]
\footnotesize
\caption{De-censoring representations}
\begin{algorithmic}[1]
\State \textbf{Input:} Auxiliary dataset $\dataa$, black-box oracle
$\enc$, observed representation $z^\star$
\State Train auxiliary model $\Maux=\enc_\text{aux}\circ\clf_\text{aux}$ on $\dataa$
\State Initialize transform model $T$, inference attack model $\Mattack$
\For{each training iteration}
\State Sample a batch of data $(x, s)$ from $\dataa$ and compute $z=\enc(x), z_\text{aux}=\enc_\text{aux}(x)$
\State Update $T$ on the batch of $(z, z_\text{aux})$ with loss $||T(z) - z_\text{aux}||_2^2$
\State Update $\Mattack$ on the batch of $(T(z), s)$ with cross-entropy loss
\EndFor
\State \Return prediction $\hat{s} = \Mattack(T(z^\star))$
\end{algorithmic}
\label{algo:decensor}
\end{algorithm}

\subsection{Re-purposing models to predict sensitive attributes}
\label{sec:transfer-attack}



To re-purpose a model\textemdash for example, to convert a
model trained for a benign task into a model that predicts a
sensitive attribute\textemdash we can use features $z_l$ in any
layer of $M$ as the feature extractor and connect a new classifier
$\clf_\text{transfer}$ to $\enc_l$.  The transferred model $\Mtransfer =
\clf_\text{transfer}\circ\enc_l$ is fine-tuned on another, small dataset
$\datat$, which in itself is not sufficient to train an accurate model
for the new task.  Utilizing features learned by $M$ on the original
$\data$, $\Mtransfer$ can achieve better results than models trained
from scratch on $\datat$.

Feasibility of model re-purposing complicates the application of policies
and regulations such as GDPR~\citep{gdpr}.  GDPR requires data processors
to disclose every purpose of data collection and obtain consent from
the users whose data was collected.  We show that, given a trained
model, it is not possible to determine\textemdash nor, consequently,
disclose or obtain user consent for\textemdash what the model has learned.
Learning per se thus cannot be a regulated ``purpose'' of data collection.
Regulators must be aware that even if the original training data has
been erased, a model can be re-purposed for a different objective,
possibly not envisioned at the time of original data collection.
We discuss this further in Section~\ref{sec:implications}.




\section{Experimental Results}

\begin{table}[t]
\footnotesize
    \centering
    \caption{\footnotesize Summary of datasets and tasks. Cramer's V
    captures statistical correlation between $y$ and $s$ (0 indicates no correlation and 1 indicates perfectly correlated).}
    \begin{tabular}{l|ccccccc}
    \toprule
    Dataset     & Health & UTKFace & FaceScrub & Places365 & Twitter & Yelp & PIPA \\
         \midrule
    Target $y$ & CCI & gender & gender & in/outdoor & age & review score & facial IDs \\
    Attribute $s$ & age & race & facial IDs & scene type & author & author & IDs together \\ 
    Cramer's V & 0.149 & 0.035 & 0.044 & 0.052 & 0.134 & 0.033 & n/a \\ \bottomrule
    \end{tabular}
\end{table}

\subsection{Datasets, tasks, and models}

\paraname{Health} is the Heritage Health dataset~\citep{healthdata} with
medical records of over 55,000 patients, binarized into 112 features
with age information removed.  The task is to predict if Charlson Index
(an estimate of patient mortality) is greater than zero; the sensitive
attribute is age (binned into 9 ranges).


\paraname{UTKFace} is a set of over 23,000 face images labeled with age,
gender, and race~\citep{utkface, Zhange2017Age}.  We rescaled them into
50$\times$50 RGB pixels.  The task is to predict gender; the sensitive
attribute is race.

\paraname{FaceScrub} is a set of face images labeled with
gender~\citep{facescrub}.  Some URLs are expired, but we were able
to download 74,000 images for 500 individuals and rescale them into
50$\times$50 RGB pixels.  The task is to predict gender; the sensitive
attribute is identity.

\paraname{Places365} is a set of 1.8 million images labeled with 365
fine-grained scene categories.  We use a subset of 73,000 images, 200
per category.  The task is to predict whether the scene is indoor or
outdoor; the sensitive attribute is the fine-grained scene label.

\paraname{Twitter} is a set of tweets from the PAN16
dataset~\citep{Rangel2016Overview} labeled with user information.
We removed tweets with fewer than 20 tokens and users with fewer than
50 tweets, yielding a dataset of over 46,000 tweets from 151 users with
an over 80,000-word vocabulary.  The task is to predict the age of the
user given a tweet; the sensitive attribute is the author's identity.

\paraname{Yelp} is a set of Yelp reviews labeled with user
identities~\citep{yelpdata}.  We removed users with fewer than 1,000
reviews and reviews with more than 200 tokens, yielding a dataset of
over 39,000 reviews from 137 users with an over 69,000-word vocabulary.
The task is to predict the review score between 1 to 5; the sensitive
attribute is the author's identity.

\paraname{PIPA} is a set of over 60,000 photos of 2,000 individuals
gathered from public Flickr photo albums~\citep{piper, Zhang2015Beyond}.
Each image can include one or more individuals.  We cropped their head
regions using the bounding boxes in the image annotations.  The task is
to predict the identity given the head region; the sensitive attribute
is whether two head regions are from the same photo.

\paragraphbe{Models.} 
For Health, we use a two-layer fully connected (FC)
neural network with 128 and 32 hidden units, respectively,
following~\citep{Xie2017ControllableIT, Moyer2018Invariant}.  For UTKFace
and FaceScrub, we use a LeNet~\citep{Lecun1998Gradient} variant: three
3$\times$3 convolutional and 2$\times$2 max-pooling layers with 16, 32,
and 64 filters, followed by two FC layers with 128 and 64 hidden units.
For Twitter and Yelp, we use text CNN~\citep{Kim2014Convolutional}.
For Places365 and PIPA, we use AlexNet~\citep{Krizhevsky2012Imagenet}
with convolutional layers pre-trained on ImageNet~\citep{Deng2009Imagenet}
and further add a 3$\times$3 convolutional layer with 128 filters and
2$\times$2 max-pooling followed by two FC layers with 128 and 64 hidden
units, respectively.

\subsection{Inferring sensitive attributes from representations}
\label{sec:inference_attack}

\paragraphbe{Setup.} We use 80\% of the data for training the target
models and 20\% for evaluation.  The size of the adversary's auxiliary
dataset is 50\% of the training data.  Success of the inference attack
is measured on the final FC layer's representation of test data.
The baseline is inference from the uncensored representation.  We also
measure the success of inference against representations censored with
$\gamma=1.0$ for adversarial training and $\beta=0.01, \lambda=0.0001$ for
information-theoretical censoring, following~\citep{Xie2017ControllableIT,
Moyer2018Invariant}.

For censoring with adversarial training, we simulate the adversary
with a two-layer FC neural network with 256 and 128 hidden units.
The number of epochs is 50 for censoring with adversarial training, 30
for the other models.  We use the Adam optimizer with the learning rate
of 0.001 and batch size of 128.  For information-theoretical censoring,
the model is based on VAE~\citep{kingma2013auto,Moyer2018Invariant}.
The encoder $q(z|x)$ has the same architecture as the CNN models with
all convolutional layers.  On top of that, the encoder outputs a mean
vector and a standard deviation vector to model the random variable $z$
with the re-parameterization trick.  The decoder $p(x|z)$ has three
de-convolution layers with up-sampling to map $z$ back to the same shape
as the input $x$.

For our inference model, we use the same architecture as the censoring
adversary.  For the PIPA inference model, which takes two representations
of faces and outputs a binary prediction of whether these faces appear
in the same photo, we use two FC layers followed by a bilinear model:
$p(s|z_1, z_2) = \sigma(h(z_1)Wh(z_2)^\top)$, where $z_1, z_2$ are the
two input representations, $h$ is the two FC layers, and $\sigma$ is
the sigmoid function.  We train the inference model for 50 epochs with
the Adam optimizer, learning rate of 0.001, and batch size of 128.


\begin{table}[t]
    \footnotesize
    \centering
    \caption{  Accuracy of inference from representations (last
    FC layer).  RAND is random guessing based on majority
    class labels; BASE is inference from the uncensored representation;
    ADV from the representation censored with adversarial training; 
    IT from the information-theoretically censored representation.
}
    \begin{tabular}{l|rrrr|rrrr}
    \toprule
            & \multicolumn{4}{c|}{Acc of predicting target $y$} & \multicolumn{4}{c}{Acc of inferring sensitive attribute $s$}  \\
            Dataset & RAND & BASE & ADV & IT & RAND & BASE & ADV & IT \\ \midrule
            Health    & 66.31 & 84.33 & 80.16 & 82.63 & 16.00 & 32.52 & 32.00 & 26.60 \\
            UTKFace  & 52.27 & 90.38 & 90.15 & 88.15 & 42.52 & 62.18 & 53.28 & 53.30 \\
            FaceScrub & 53.53 & 98.77 & 97.90 & 97.66 &  1.42 & 33.65 & 30.23 & 10.61 \\
            Places365  & 56.16 & 91.41 & 90.84 & 89.82 &  1.37 & 31.03 & 12.56 & 2.29 \\
            Twitter  & 45.17 & 76.22 & 57.97 & n/a     & 6.93 & 38.46 & 34.27 & n/a \\
            Yelp   & 42.56 & 57.81 & 56.79 & n/a     & 15.88 & 33.09 & 27.32 & n/a \\
            PIPA  & 7.67 & 77.34 & 52.02 & 29.64 & 68.50 & 87.95 & 69.96 & 82.02 \\ \bottomrule
    \end{tabular}
    \label{tab:inference_attack}
\end{table}

\paragraphbe{Results.} Table~\ref{tab:inference_attack} reports the
results.  When representations are not censored, accuracy of inference
from the last-layer representations is much higher than random guessing
for all tasks, which means \textbf{models overlearn even in the higher,
task-specific layers}.  When representations are censored with adversarial
training, accuracy drops for both the main and inference tasks.  Accuracy
of inference is much higher than in~\citep{Xie2017ControllableIT}.  The
latter uses logistic regression, which is weaker than the training-time
censoring-adversary network, whereas we use the same architecture for
both the training-time and post-hoc adversaries.  Information-theoretical
censoring reduces accuracy of inference, but also damages main-task
accuracy more than adversarial training for almost all models.

\label{sec:notrepresented}

\textbf{Overlearning can cause a model to recognize even the sensitive
attributes that are not represented in the training dataset}.  Such
attributes cannot be censored using any known technique.  We trained a
UTKFace gender classifier on datasets where all faces are of the same
race.  We then applied this model to test images with four races (White,
Black, Asian, Indian) and attempted to infer the race attribute from
the model's representations.  Inference accuracy is 61.95\%, 61.99\%,
60.85\% and 60.81\% for models trained only on, respectively, White,
Black, Asian, and Indian images\textemdash almost as good as the 62.18\%
baseline and much higher than random guessing (42.52\%).


\begin{figure}[t]
\newcommand{\plotwidth}{0.28\textwidth}
\newcommand{\plotheight}{0.22\textwidth}

\begin{tikzpicture}
\begin{axis}[title style={yshift=-1.5ex}, title=\textbf{FaceScrub}, xlabel=$\gamma$, xmin=1, xmax=6, ylabel=\textbf{ADV} \\ Relative acc
, ylabel style={align=center}, height=\plotheight,
width=\plotwidth, grid=both, name=ax1,
legend pos=south west]
\pgfplotstableread{figs/facescrub_gamma.txt}\mydata;
\addplot[thick, mark=o, color=blue] table
[x expr=\thisrow{gamma},y expr=\thisrow{task} - 98.77] {\mydata};
\addplot[thick, mark=triangle, color=red] table
[x expr=\thisrow{gamma},y expr=\thisrow{attack} - 33.65] {\mydata};
\end{axis}
\begin{axis}[title style={yshift=-1.5ex}, title=\textbf{UTKFace}, xlabel=$\gamma$,
xmin=0.1, xmax=2,  height=\plotheight,
width=\plotwidth, grid=both, name=ax2,at={(ax1.south east)}, xshift=1cm]
\pgfplotstableread{figs/utk_gamma.txt}\mydata;
\addplot[thick, mark=o, color=blue] table
[x expr=\thisrow{gamma},y expr=\thisrow{task} - 90.38] {\mydata};
\addplot[thick, mark=triangle, color=red] table
[x expr=\thisrow{gamma},y expr=\thisrow{attack} - 62.18] {\mydata};
\end{axis}
\begin{axis}[title style={yshift=-1.5ex}, title=\textbf{Twitter}, xlabel=$\gamma$, 
xmin=0.2, xmax=1, width=\plotwidth,  height=\plotheight, grid=both, 
at={(ax2.south east)}, name=ax3, xshift=1cm]
\pgfplotstableread{figs/twitter_gamma.txt}\mydata;
\addplot[thick, mark=o, color=blue] table
[x expr=\thisrow{gamma},y expr=\thisrow{task} - 76.22] {\mydata};
\addplot[thick, mark=triangle, color=red] table
[x expr=\thisrow{gamma},y expr=\thisrow{attack} - 38.46] {\mydata};
\end{axis}
\begin{axis}[title style={yshift=-1.5ex}, title=\textbf{Yelp}, xlabel=$\gamma$, 
xmin=0.1, xmax=2, width=\plotwidth, height=\plotheight, grid=both, 
at={(ax3.south east)}, xshift=1cm]
\pgfplotstableread{figs/yelp_gamma.txt}\mydata;
\addplot[thick, mark=o, color=blue] table
[x expr=\thisrow{gamma},y expr=\thisrow{task} - 57.81] {\mydata};
\addplot[thick, mark=triangle, color=red] table
[x expr=\thisrow{gamma},y expr=\thisrow{attack} - 38.46] {\mydata};
\end{axis}
\end{tikzpicture}

\begin{tikzpicture}
\begin{axis}[title style={yshift=-1.5ex}, title=\textbf{Health}, xlabel=$\beta\cdot 10^2$,ylabel=\textbf{IT} \\ Relative acc, 
ylabel style={align=center}, xmin=1, xmax=8,
xtick=data,
xticklabels={1,2,4,6,8},
width=\plotwidth, height=\plotheight,
grid=both, name=ax1]
\pgfplotstableread{figs/health_beta.txt}\mydata;
\addplot[thick, mark=o, color=blue] table
[x expr=\thisrow{beta} * 100,y expr=\thisrow{task} - 84.33] {\mydata};
\addplot[thick, mark=triangle, color=red] table
[x expr=\thisrow{beta} * 100,y expr=\thisrow{attack} - 32.52] {\mydata};
\end{axis}
\begin{axis}[title style={yshift=-1.5ex}, title=\textbf{UTKFace}, xlabel=$\beta\cdot 10^2$,
xmin=1, xmax=2,
width=\plotwidth, height=\plotheight, grid=both,
legend style = {at = {(0.5,-0.8)}, legend columns=2, anchor=south},
at={(ax1.south east)}, xshift=1cm, name=ax2]
\pgfplotstableread{figs/utk_beta.txt}\mydata;
\addplot[thick, mark=o, color=blue] table
[x expr=\thisrow{beta} * 100,y expr=\thisrow{task} - 90.38] {\mydata};
\addplot[thick, mark=triangle, color=red] table
[x expr=\thisrow{beta} * 100,y expr=\thisrow{attack} - 62.18] {\mydata};
\end{axis}

\begin{axis}[title style={yshift=-1.5ex}, title=\textbf{FaceScrub}, xlabel=$\beta\cdot 10^2$,
xmin=0.5, xmax=1.5,
xtick=data,
width=\plotwidth, height=\plotheight, grid=both,
legend style = {at = {(0.5,-0.8)}, legend columns=2, anchor=south},
at={(ax2.south east)}, xshift=1cm, name=ax3]
\pgfplotstableread{figs/facescrub_beta.txt}\mydata;
\addplot[thick, mark=o, color=blue] table
[x expr=\thisrow{beta}*100,y expr=\thisrow{task} - 98.77] {\mydata};
\addplot[thick, mark=triangle, color=red] table
[x expr=\thisrow{beta}*100,y expr=\thisrow{attack} - 33.65] {\mydata};
\end{axis}
\begin{axis}[title style={yshift=-1.5ex}, title=\textbf{Places365}, xlabel=$\beta\cdot 10^2$,  xmin=1, xmax=5,
width=\plotwidth, height=\plotheight, grid=both,
at={(ax3.south east)}, xshift=1cm,]
\pgfplotstableread{figs/place_beta.txt}\mydata;
\addplot[thick, mark=o, color=blue] table
[x expr=\thisrow{beta}*100,y expr=\thisrow{task} - 91.41] {\mydata};
\addplot[thick, mark=triangle, color=red] table
[x expr=\thisrow{beta}*100,y expr=\thisrow{attack} - 31.03] {\mydata};
\end{axis}
\end{tikzpicture}
\\
\begin{tikzpicture}
\begin{axis}[title style={yshift=-1.5ex}, title=\textbf{Health}, xlabel=$\lambda$, ylabel=\textbf{IT} \\ Relative acc
, ylabel style={align=center}, xmin=0.005, xmax=0.5, xmode=log,
xtick=data, xticklabels={,0.01, , 0.1, 0.5},
width=\plotwidth, height=\plotheight, grid=both,
 name=ax1]
\pgfplotstableread{figs/health_lambda.txt}\mydata;
\addplot[thick, mark=o, color=blue] table
[x expr=\thisrow{lambda},y expr=\thisrow{task} - 84.33] {\mydata};
\addplot[thick, mark=triangle, color=red] table
[x expr=\thisrow{lambda},y expr=\thisrow{attack} - 32.52] {\mydata};
\end{axis}

\begin{axis}[title style={yshift=-1.5ex}, title=\textbf{UTKFace}, xlabel=$\lambda\cdot 10^3$,  xmin=0.1, xmax=1,
width=\plotwidth, height=\plotheight, grid=both,
legend style = {at = {(0.5,-0.6)}, legend columns=2, anchor=south},
at={(ax1.south east)}, xshift=1cm, name=ax2]
\pgfplotstableread{figs/utk_lambda.txt}\mydata;
\addplot[thick, mark=o, color=blue] table
[x expr=\thisrow{lambda} * 1000,y expr=\thisrow{task} - 90.38] {\mydata};
\addplot[thick, mark=triangle, color=red] table
[x expr=\thisrow{lambda} * 1000,y expr=\thisrow{attack} - 62.18] {\mydata};
\end{axis}

\begin{axis}[title style={yshift=-1.5ex}, title=\textbf{FaceScrub}, xlabel=$\lambda$,  xmin=0.00001, xmax=0.1, xmode=log,
width=\plotwidth, height=\plotheight, grid=both,
legend style = {at = {(0.5,-0.6)}, legend columns=2, anchor=south},
at={(ax2.south east)}, xshift=1cm, name=ax3]
\pgfplotstableread{figs/facescrub_lambda.txt}\mydata;
\addplot[thick, mark=o, color=blue] table
[x expr=\thisrow{lambda},y expr=\thisrow{task} - 98.77] {\mydata};
\addplot[thick, mark=triangle, color=red] table
[x expr=\thisrow{lambda},y expr=\thisrow{attack} - 33.65] {\mydata};
\end{axis}

\begin{axis}[title style={yshift=-1.5ex}, title=\textbf{Places365}, xlabel=$\lambda$, xmin=0.00001, xmax=0.001, xmode=log,
xtick=data, 
xticklabels={$10^{-5}$, , $10^{-4}$, , $10^{-3}$},
width=\plotwidth, height=\plotheight, grid=both,
at={(ax3.south east)}, xshift=1cm]
\pgfplotstableread{figs/place_lambda.txt}\mydata;
\addplot[thick, mark=o, color=blue] table
[x expr=\thisrow{lambda},y expr=\thisrow{task} - 91.41] {\mydata};
\addplot[thick, mark=triangle, color=red] table
[x expr=\thisrow{lambda},y expr=\thisrow{attack} - 31.03] {\mydata};
\end{axis}
\end{tikzpicture}

\caption{ 
Reduction in accuracy due to censoring.  Blue lines are the main task,
red lines are the inference of sensitive attributes.  First row is
adversarial training with different $\gamma$ values; second and third
row is information-theoretical censoring with different $\beta$ and
$\lambda$ values respectively.} 
\label{fig:gamma} 
\end{figure}
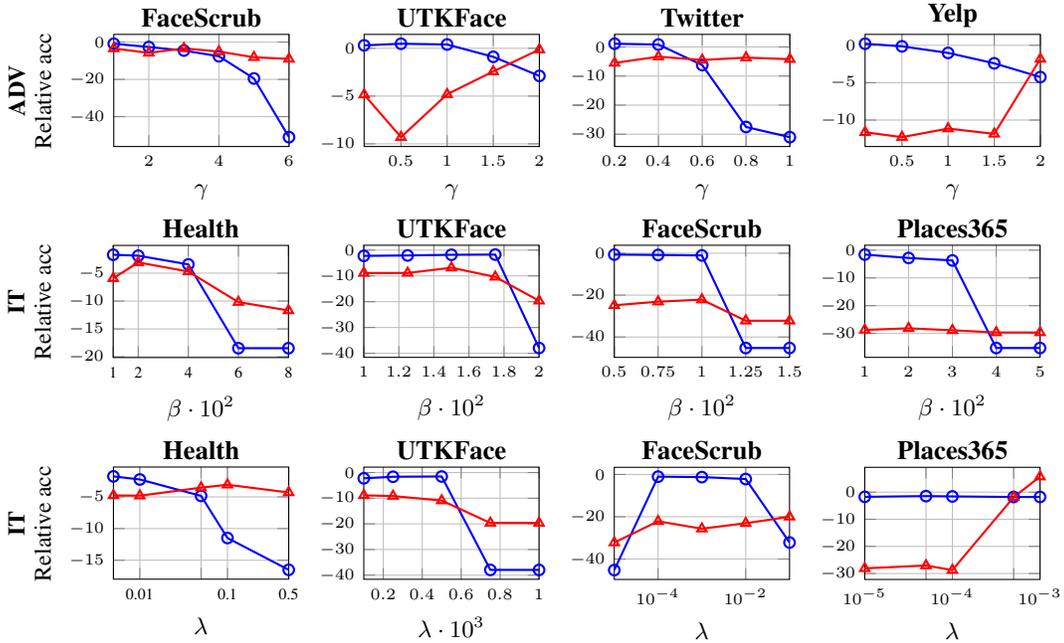


\paragraphbe{Effect of censoring strength.} 
Fig.~\ref{fig:gamma} shows that stronger censoring does not help.
On FaceScrub and Twitter with adversarial training, increasing $\gamma$
damages the model's accuracy on the main task, while accuracy of inference
decreases slightly or remains the same.  For UTKFace and Yelp, increasing
$\gamma$ \emph{improves} accuracy of inference.  This may indicate that
the simulated ``adversary'' during adversarial training overpowers the
optimization process and censoring defeats itself.

For all models with information-theoretical censoring, increasing
$\beta$ reduces the accuracy of inference but can lead to the model
not converging on its main task.  Increasing $\lambda$ results in
the model not converging on the main task, without affecting the
accuracy of inference, on Health, UTKFace and FaceScrub.  This seems
to contradict the censoring objective, but the reconstruction loss
in Equation~\ref{eq:itcensor} dominates the other loss terms, which
leads to poor divergence between conditional $q(z|x)$ and $q(z)$, i.e.,
information about $x$ is still retained in $z$.

\begin{table}[t]
\footnotesize
    \centering
    \caption{Improving inference accuracy
    with de-censoring. $\delta$ is the increase from
    Table~\ref{tab:inference_attack}.}
    \begin{tabular}{r|rrrrrr}
    \toprule
    Dataset & Health & UTKFace & FaceScrub & Places365 & Twitter & Yelp \\ \midrule
    ADV $+\delta$ & 32.55 +0.55 & 59.38 +6.10 & 40.37 +12.24 &  19.71 +7.15 & 36.55 +2.22 & 31.36 +4.04  \\
    IT  $+\delta$ & 27.05 +0.45 & 54.31 +1.01 & 16.40 \hspace{.5em}+5.79 & 3.10 +0.81 & n/a & n/a \\ \bottomrule
    \end{tabular}
    \label{tab:decensor}
\end{table}

\paragraphbe{De-censoring.} 
As described in Section~\ref{sec:decensor}, we developed a new technique
to transform censored representations to make inference easier.
We first train an auxiliary model on $\dataa$ to predict the sensitive
attribute from representations, using the same architecture as in the
baseline models.  The resulting uncensored representations from the last
convolutional layer are the target for the de-censoring transformations.
We use a single-layer fully connected neural network as the transformer
and set the number of hidden units to the dimension of the uncensored
representation.  The inference model operates on top of the transformer
network, with the same hyper-parameters as before.

Table~\ref{tab:decensor} shows that de-censoring significantly boosts
the accuracy of inference from representations censored with adversarial
training.  The boost is smaller against information-theoretical censoring
because its objective not only censors $z$ with $I(z, s)$, but also
forgets $x$ with $I(x, z)$.  On the Health task, there is not much
difference since the baseline attack is already similar to the attack
on censored representations, leaving little room for improvement.

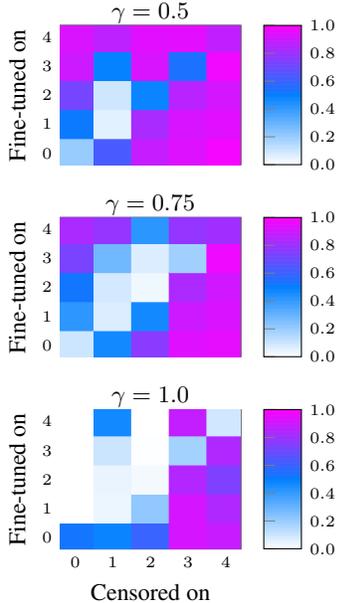
\begin{table}[t]
    \centering
    \footnotesize
\caption{Adversarial re-purposing.  The values are differences between
the accuracy of predicting sensitive attributes using a re-purposed
model vs.\ a model trained from scratch.}
    \begin{tabular}{l|rrrrrrr}
    \toprule
    $|\datat| / |\data|$     & Health & UTKFace & FaceScrub & Places365 & Twitter & Yelp & PIPA \\
    \midrule
    0.02 & -0.57 & 4.72 & 7.01 & 4.42 & 12.99 & 5.57 & 1.33 \\
    0.04 & 0.22 & 2.70 & 15.07 & 2.14 & 10.87 & 3.60 & 2.41 \\
    0.06 & -1.21 & 2.83 & 7.02 & 2.06 & 10.51 & 8.45 & 6.50 \\
    0.08 & -0.99 & 0.25 & 11.80 & 3.39 & 9.57 & 0.33 &  4.93 \\
    0.10 & 0.35 & 2.24 & 9.43 & 2.86   & 7.30 & 2.1 & 5.89 \\ \bottomrule
    \end{tabular}
    \label{tab:adv_repurpose}
\end{table}

In summary, these results demonstrate that information about sensitive
attributes unintentionally captured by the overlearned representations
cannot be suppressed by censoring.

\subsection{Re-purposing models to predict sensitive attributes}

To demonstrate that \textbf{overlearned representations can be picked
up by a small set of unseen data to create a model for predicting
sensitive attributes}, we re-purpose uncensored baseline models from
Section~\ref{sec:inference_attack} by fine-tuning them on a small
($2-10\%$ of $\data$) set $\datat$ and compare with the models
trained from scratch on $\datat$.  We fine-tune all models for 50
epochs with batch size of 32; the other hyper-parameters are as in
Section~\ref{sec:inference_attack}.  For all CNN models, we use the
trained convolutional layers as the feature extractor and randomly
initialize the other layers.  Table~\ref{tab:adv_repurpose} shows that
the re-purposed models always outperform those trained from scratch.
FaceScrub and Twitter exhibit the biggest gain.



\begin{table}[t]
    \footnotesize
    \caption{The effect of censoring on adversarial re-purposing for
    FaceScrub with $\gamma=0.5, 0.75, 1.0$.  $\delta_A$ is the difference
    in the original-task accuracy (second column) between uncensored
    and censored models; $\delta_B$ is the difference in the accuracy
    of inferring the sensitive attribute (columns 3 to 7) between the
    models re-purposed from different layers and the model trained from
    scratch.  Negative values mean reduced accuracy.  Heatmaps on the
    right are linear CKA similarities between censored and uncensored
    representations.  Numbers 0 through 4 represent layers conv1, conv2,
    conv3, fc4, and fc5.  For each model censored at layer $i$ (x-axis),
    we measure similarity between the censored and uncensored models at
    layer $j$ (y-axis).}
    \label{tab:transfer_censor}
\begin{minipage}{0.68\linewidth}
    \centering
    \setlength{\tabcolsep}{5pt}
    \begin{tabular}{l|r|rrrrr}
    \toprule 
         Censored on&  & \multicolumn{5}{c}{$\delta_B$ when transferred from}  \\ 
         $\gamma=0.5$ & $\delta_A$  & conv1 & conv2 & conv3 & fc4 & fc5  \\ \midrule
        conv1 & -1.66 & -6.42 & -4.09 & -1.65 & 0.46 & -3.87 \\
        conv2 & -2.87 & 0.95 & -1.77 & -2.88 & -1.53 & -2.22 \\
        conv3 & -0.64 & 1.49 & 1.49 & 0.67 & -0.48 & -1.38 \\
        fc4 & -0.16 & 2.03 & 5.16 & 6.73 & 6.12 & 0.54 \\
        fc5 & 0.05 & 1.52 & 4.53 & 7.42 & 6.14 & 4.53 \\\midrule
         $\gamma=0.75$  & \multicolumn{6}{c}{}  \\ \midrule
        conv1 & -4.48 & -7.33 & -5.01 & -1.51 & -7.99 & -7.82 \\
        conv2 & -6.02 & 0.44 & -7.04 & -5.46 & -5.94 & -5.82 \\
        conv3 & -1.90 & 1.32 & 1.37 & 1.88 & 0.74 & -0.67 \\
        fc4 & 0.01 & 3.65 & 4.56 & 5.11 & 4.44 & 0.91 \\
        fc5 & -0.74 & 1.54 & 3.61 & 6.75 & 7.18 & 4.99 \\ \midrule
        $\gamma=1$ & \multicolumn{6}{c}{} \\ \midrule
        conv1 & -45.25 & -7.36 & -3.93 & -2.75 & -4.37 & -2.91 \\
        conv2 & -20.30 & -3.28 & -5.27 & -7.03 & -6.38 & -5.54 \\
        conv3 & -45.20 & -2.13 & -3.06 & -4.48 & -4.05 & -5.18 \\
        fc4 & -0.52 & 1.73 & 5.19 & 4.80 & 5.83 & 1.84 \\
        fc5 & -0.86 & 1.56 & 3.55 & 5.59 & 5.14 & 1.97 \\\bottomrule
\end{tabular}
\end{minipage}
\hfill
\begin{minipage}{0.3\linewidth}
\centering
\begin{tikzpicture}
\begin{groupplot}[
    group style={
        group name=my plots,
        group size=1 by 3,
        xlabels at=edge bottom,
        xticklabels at=edge bottom,
        vertical sep=20pt,
    },
    width=0.95\textwidth,
    title style={yshift=-2ex},
    xtick={0, 1, 2, 3, 4},
    ytick={0, 1, 2, 3, 4},
    ylabel=Fine-tuned on,
    tickpos=left,
    point meta min=0, point meta max=1, colormap/cool,
    colorbar, colorbar style={yticklabel style={ /pgf/number format/.cd, fixed, precision=1, fixed zerofill,}, },
]
\pgfplotstableread{figs/facescrub_adv_heatmap.txt}\mydata;
\nextgroupplot[title={$\gamma=0.5$}]
\addplot [matrix plot*, mesh/cols=5, point meta=explicit] table [meta=g1] {\mydata};
\nextgroupplot[title={$\gamma=0.75$}]
\addplot [matrix plot*, mesh/cols=5, point meta=explicit] table [meta=g2] {\mydata};
\nextgroupplot[title={$\gamma=1.0$}, xlabel=Censored on]
\addplot [matrix plot*, mesh/cols=5, point meta=explicit] table [meta=g3] {\mydata};
\end{groupplot}
\end{tikzpicture}

\end{minipage}
\end{table}

\paragraphbe{Effect of censoring.} 
Previous work only censored the highest layer of the models.  Model
re-purposing can use any layer of the model for transfer learning.
Therefore, to prevent re-purposing, inner layers must be censored, too.
We perform the first study of inner-layers censoring and measure its
effect on both the original and re-purposed tasks.  We use FaceScrub
for this experiment and apply adversarial training to every layer with
different strengths ($\gamma=0.5, 0.75, 1.0$).

Table~\ref{tab:transfer_censor} summarizes the results.  Censoring lower
layers (conv1 to conv3) blocks adversarial re-purposing, at the cost of
reducing the model's accuracy on its original task.  Hyper-parameters
must be tuned carefully, e.g. when $\gamma=1$, there is a huge drop in
the original-task accuracy.

To further investigate how censoring in one layer affects the
representations learned across all layers, we measure per-layer
similarity between censored and uncensored models using CKA, linear
centered kernel alignment~\citep{Kornblith2019Similarity}\textemdash see
Table~\ref{tab:transfer_censor}.  When censoring is applied to a specific
layer, similarity for that layer is the smallest (values on the diagonal).
When censoring lower layers with moderate strength ($\gamma=0.5$ or
$0.75$), similarity between higher layers is still strong; when censoring
higher layers, similarity between lower layers is strong.  Therefore,
censoring can block adversarial re-purposing from a specific layer, but
the adversary can still re-purpose representations in the other layer(s)
to obtain an accurate model for predicting sensitive attributes.


\subsection{When, where, and why overlearning happens}

To investigate when (during training) and where
(in which layer) the models overlearn, we use linear CKA
similarity~\citep{Kornblith2019Similarity} to compare the representations
at different epochs of training between models trained for the original
task (A) and models trained to predict a sensitive attribute (B).
We use UTKFace and FaceScrub for these experiments.

Fig.~\ref{fig:epoch_sim} shows that lower layers of models
A and B learn very similar features.  This was observed
in~\citep{Kornblith2019Similarity} for CIFAR-10 and CIFAR-100 models,
but those tasks are closely related.  In our case, the tasks are
entirely different and B reveals the sensitive attribute while A
does not.  The similar low-level features are learned very early during
training.  There is little similarity between the low-level features of
A and high-level features of B (and vice versa), matching intuition.
Interestingly, on FaceScrub even the high-level features are similar
between A and B.

\newcommand{\cellwidth}{0.26\textwidth}
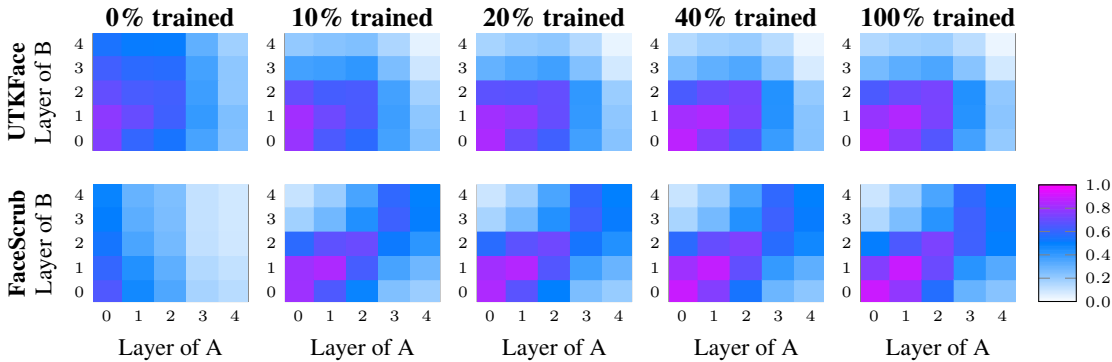
\begin{figure}[t]
\begin{tikzpicture}
\begin{axis}[
title style={yshift=-1.5ex}, title=\textbf{0\% trained}, 
width=\cellwidth,  ylabel style={align=center}, ylabel=\textbf{UTKFace} \\ Layer of B,
xticklabels={,,}, ytick={0,1,2,3,4},
point meta min=0, point meta max=1, colormap/cool, name=ax1]
\pgfplotstableread{figs/utk_epoch_heatmap.txt}\mydata;
\addplot [matrix plot*, mesh/rows=5, mesh/cols=5, point meta=explicit] table [x=x,y=y,meta=e0] {\mydata};
\end{axis}
\begin{axis}[
title style={yshift=-1.5ex}, title=\textbf{10\% trained},
 width=\cellwidth,
xticklabels={,,}, ytick={0,1,2,3,4},
point meta min=0, point meta max=1, colormap/cool, name=ax2 ,at={(ax1.south east)}, xshift=0.5cm]
\pgfplotstableread{figs/utk_epoch_heatmap.txt}\mydata;
\addplot [matrix plot*,  mesh/rows=5, mesh/cols=5,  point meta=explicit] table [x=x,y=y,meta=e3] {\mydata};
\end{axis}
\begin{axis}[
title style={yshift=-1.5ex}, title=\textbf{20\% trained},
width=\cellwidth,
xticklabels={,,}, ytick={0,1,2,3,4},
point meta min=0, point meta max=1, colormap/cool, name=ax3, at={(ax2.south east)}, xshift=0.5cm]
\pgfplotstableread{figs/utk_epoch_heatmap.txt}\mydata;
\addplot [matrix plot*,  mesh/rows=5, mesh/cols=5,  point meta=explicit] table [x=x,y=y,meta=e6] {\mydata};
\end{axis}
\begin{axis}[
title style={yshift=-1.5ex}, title=\textbf{40\% trained},
 width=\cellwidth,
xticklabels={,,}, ytick={0,1,2,3,4},
point meta min=0, point meta max=1, colormap/cool, name=ax4, at={(ax3.south east)}, xshift=0.5cm]
\pgfplotstableread{figs/utk_epoch_heatmap.txt}\mydata;
\addplot [matrix plot*,  mesh/rows=5, mesh/cols=5,  point meta=explicit] table [x=x,y=y,meta=e12] {\mydata};
\end{axis}
\begin{axis}[
title style={yshift=-1.5ex}, title=\textbf{100\% trained},
width=\cellwidth,
xticklabels={,,}, ytick={0,1,2,3,4},
point meta min=0, point meta max=1, colormap/cool,
at={(ax4.south east)}, xshift=0.5cm]
\pgfplotstableread{figs/utk_epoch_heatmap.txt}\mydata;
\addplot [matrix plot*,  mesh/rows=5, mesh/cols=5,  point meta=explicit] table [x=x,y=y,meta=e30] {\mydata};
\end{axis}
\end{tikzpicture}

\begin{tikzpicture}
\begin{axis}[
title style={yshift=-1.5ex}, 
xlabel=Layer of A, ylabel style={align=center}, ylabel=\textbf{FaceScrub} \\ Layer of B,  width=\cellwidth,
xtick={0,1,2,3,4}, ytick={0,1,2,3,4},
point meta min=0, point meta max=1, colormap/cool, name=ax1]
\pgfplotstableread{figs/facescrub_epoch_heatmap.txt}\mydata;
\addplot [matrix plot*, mesh/rows=5, mesh/cols=5, point meta=explicit] table [x=x,y=y,meta=e0] {\mydata};
\end{axis}
\begin{axis}[
title style={yshift=-1.5ex}, 
xlabel=Layer of A,  width=\cellwidth,
xtick={0,1,2,3,4}, ytick={0,1,2,3,4},
point meta min=0, point meta max=1, colormap/cool, name=ax2 ,at={(ax1.south east)}, xshift=0.5cm]
\pgfplotstableread{figs/facescrub_epoch_heatmap.txt}\mydata;
\addplot [matrix plot*,  mesh/rows=5, mesh/cols=5,  point meta=explicit] table [x=x,y=y,meta=e3] {\mydata};
\end{axis}
\begin{axis}[
title style={yshift=-1.5ex}, 
xlabel=Layer of A,  width=\cellwidth,
xtick={0,1,2,3,4}, ytick={0,1,2,3,4},
point meta min=0, point meta max=1, colormap/cool, name=ax3, at={(ax2.south east)}, xshift=0.5cm]
\pgfplotstableread{figs/facescrub_epoch_heatmap.txt}\mydata;
\addplot [matrix plot*,  mesh/rows=5, mesh/cols=5,  point meta=explicit] table [x=x,y=y,meta=e6] {\mydata};
\end{axis}
\begin{axis}[
title style={yshift=-1.5ex}, 
xlabel=Layer of A,  width=\cellwidth,
xtick={0,1,2,3,4}, ytick={0,1,2,3,4},
point meta min=0, point meta max=1, colormap/cool, name=ax4, at={(ax3.south east)}, xshift=0.5cm]
\pgfplotstableread{figs/facescrub_epoch_heatmap.txt}\mydata;
\addplot [matrix plot*,  mesh/rows=5, mesh/cols=5,  point meta=explicit] table [x=x,y=y,meta=e12] {\mydata};
\end{axis}
\begin{axis}[
title style={yshift=-1.5ex}, 
xlabel=Layer of A,  width=\cellwidth,
xtick={0,1,2,3,4}, ytick={0,1,2,3,4},
colorbar, 
colorbar style={
yticklabel style={ /pgf/number format/.cd, fixed, precision=1, fixed zerofill,}},
point meta min=0, point meta max=1, colormap/cool,
at={(ax4.south east)}, xshift=0.5cm]
\pgfplotstableread{figs/facescrub_epoch_heatmap.txt}\mydata;
\addplot [matrix plot*,  mesh/rows=5, mesh/cols=5,  point meta=explicit] table [x=x,y=y,meta=e30] {\mydata};
\end{axis}
\end{tikzpicture}
\caption{Pairwise similarities of layer representations
between models for the original task (A) and for predicting a sensitive
attribute (B).  Numbers 0 through 4 denote layers conv1, conv2, conv3,
fc4 and fc5.}
\label{fig:epoch_sim}
\end{figure}
\begin{figure}[t]
\centering
\begin{tikzpicture}
\pgfplotstableread{figs/facescrub50_epoch_sim.txt}\mydataa;
\pgfplotstableread{figs/facescrub500_epoch_sim.txt}\mydatab;

\begin{axis}[title style={yshift=-1.5ex}, title=\textbf{Conv1}, 
xlabel=Epoch, xmin=10, xmax=30, ylabel= Similarity to \\ Random Weights, ylabel style={align=center}, width=0.35\textwidth, grid=both, name=ax1,
legend pos=south west]
\addplot[thick, mark=o, color=blue] table
[x expr=\thisrow{epoch},y expr=\thisrow{pool1}] {\mydataa};
\addplot[thick, mark=triangle, color=red] table
[x expr=\thisrow{epoch},y expr=\thisrow{pool1}] {\mydatab};
\end{axis}

\begin{axis}[title style={yshift=-1.5ex}, title=\textbf{Conv2}, 
xlabel=Epoch, xmin=10, xmax=30, width=0.35\textwidth, grid=both,  
name=ax2 ,at={(ax1.south east)}, xshift=1.0cm,
legend pos=north east]
\addplot[thick, mark=o, color=blue] table
[x expr=\thisrow{epoch},y expr=\thisrow{pool2}] {\mydataa};
\addplot[thick, mark=triangle, color=red] table
[x expr=\thisrow{epoch},y expr=\thisrow{pool2}] {\mydatab};
\end{axis}

\begin{axis}[title style={yshift=-1.5ex}, title=\textbf{Conv3}, 
xlabel=Epoch, xmin=10, xmax=30, width=0.35\textwidth, grid=both,  
name=ax3 ,at={(ax2.south east)}, xshift=1.0cm,
legend pos=north east]
\addplot[thick, mark=o, color=blue] table
[x expr=\thisrow{epoch},y expr=\thisrow{pool3}] {\mydataa};
\addlegendentry{50 IDs};
\addplot[thick, mark=triangle, color=red] table
[x expr=\thisrow{epoch},y expr=\thisrow{pool3}] {\mydatab};
\addlegendentry{500 IDs};
\end{axis}
\end{tikzpicture}
\caption{ 
Similarity of layer representations of a partially trained
gender classifier to a randomly initialized model before training. 
Models are trained on FaceScrub using 50 IDs (blue line) and 500 IDs (red line).}
\label{fig:simtorandom}
\end{figure}
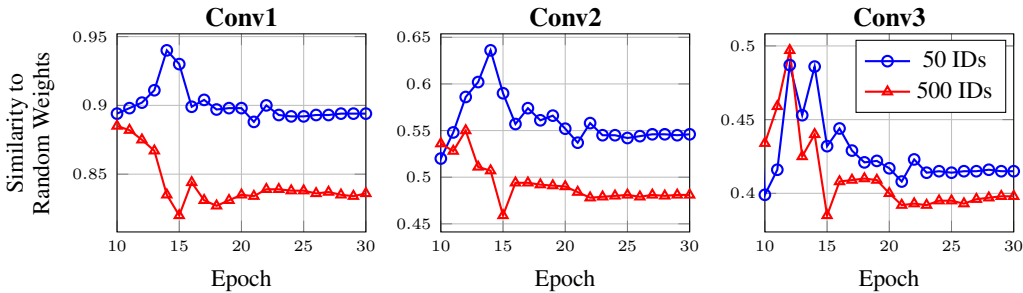

We conjecture that one of the reasons for overlearning is structural
complexity of the data.  Previous work theoretically showed that
over-parameterized neural networks favor simple solutions on structured
data when optimized with SGD, where structure is quantified as the
number of distributions (e.g., images from different identities) within
each class in the target task~\citep{li2018learning}, i.e., the fewer
distributions, the more structured the data.  For data generated from
more complicated distributions, networks learn more complex solutions,
leading to the emergence of features that are much more general than
the learning objective and, consequently, overlearning.

Fig.~\ref{fig:simtorandom} shows that the representations of a gender
classifier trained on the faces from 50 individuals are closer to the
random initialization than the representations trained on the faces from
500 individuals (the hyper-parameters and the total number of training
examples are the same in both cases).  More complex training data thus
results in more complex representations for the same objective.

\section{Related Work}


Prior work studied transferability of representations only between
closely related tasks.  Transferability of features between ImageNet
models decreases as the distance between the base and target tasks
grows~\citep{Yosinski2014How}, and performance of tasks is correlated
to their distance from the source task~\citep{Azizpour2015Generic}.
CNN models trained to distinguish coarse classes also distinguish their
subsets~\citep{Huh2016Makes}.  By contrast, we show that models trained
for simple tasks implicitly learn privacy-sensitive concepts unrelated
to the labels of the original task.  Other than an anecdotal mention
in the acknowledgments paragraph of~\citep{Kim2017Interpretability}
that logit-layer activations leak non-label concepts, this phenomenon
has never been described in the research literature.


Gradient updates revealed by participants in distributed learning leak
information about individual training batches that is uncorrelated
with the learning objective~\citep{Melis2019Exploiting}.  We show that
overlearning is a generic problem in (fully trained) models, helping
explain these observations.


There is a large body of research on learning disentangled
representations~\citep{bengio2013representation,
locatello2019challenging}.  The goal is to separate the
underlying explanatory factors in the representation so
that it contains all information about the input in an
interpretable structure.  State-of-the-art approaches
use variational autoencoders~\citep{kingma2013auto} and
their variants to learn disentangled representations in an
unsupervised fashion~\citep{higgins2017beta, kumar2017variational,
kim2018disentangling, chen2018isolating}.  By contrast, overlearning
means that representations learned during supervised training for one
task implicitly and automatically enable another task\textemdash without
disentangling the representation on purpose during training.



Work on censoring representations aims to suppress sensitive demographic
attributes and identities in the model's output for fairness and privacy.
Techniques include adversarial training~\citep{Edwards2016CensoringRW},
which has been applied to census and health
records~\citep{Xie2017ControllableIT}, text~\citep{Li2018Towards,
Coavoux2018Privacy, Elazar2018Adversarial}, images~\citep{Hamm2017Minimax}
and sensor data of wearables~\citep{Iwasawa2017Privacy}.  An alternative
approach is to minimize mutual information between the representation
and the sensitive attribute~\citep{Moyer2018Invariant, Osia2018Deep}.
Neither approach can prevent overlearning, except at the cost of
destroying the model's accuracy.  Furthermore, these techniques cannot
censor attributes that are not represented in the training data.  We show
that overlearned models recognize such attributes, too.


\section{Conclusions}
\label{sec:implications}

We demonstrated that models trained for seemingly simple tasks implicitly
learn concepts that are not represented in the objective function.
In particular, they learn to recognize sensitive attributes, such as
race and identity, that are statistically orthogonal to the objective.
The failure of censoring to suppress these attributes and the similarity
of learned representations across uncorrelated tasks suggest that
overlearning may be intrinsic, i.e., learning for some objectives may not
be possible without recognizing generic low-level features that enable
other tasks, including inference of sensitive attributes.  For example,
there may not exist a set of features that enables a model to accurately
determine the gender of a face but not its race or identity.

This is a challenge for regulations such as GDPR that aim to control the
purposes and uses of machine learning technologies.  To protect privacy
and ensure certain forms of fairness, users and regulators may desire
that models not learn some features and attributes.  If overlearning is
intrinsic, it may not be technically possible to enumerate, let alone
control, what models are learning.  Therefore, regulators should focus
on ensuring that models are applied in a way that respects privacy and
fairness, while acknowledging that they may still recognize and use
sensitive attributes.

\paragraphbe{Acknowledgments.}
This research was supported in part by NSF grants 1611770, 1704296, and
1916717, the generosity of Eric and Wendy Schmidt by recommendation of
the Schmidt Futures program, and a Google Faculty Research Award.

\bibliographystyle{iclr2020_conference}
\bibliography{citation}

\end{document}